\documentclass[sigconf]{acmart}

\AtBeginDocument{%
  \providecommand\BibTeX{{%
    \normalfont B\kern-0.5em{\scshape i\kern-0.25em b}\kern-0.8em\TeX}}}

\copyrightyear{2023}
\acmYear{2023}
\setcopyright{acmlicensed}
\acmConference[MM '23] {Proceedings of the 31st ACM International Conference on Multimedia}{October 29--November 3, 2023}{Ottawa, ON, Canada.}
\acmBooktitle{Proceedings of the 31st ACM International Conference on Multimedia (MM '23), October 29--November 3, 2023, Ottawa, ON, Canada}
\acmPrice{15.00}
\acmISBN{979-8-4007-0108-5/23/10}
\acmDOI{10.1145/3581783.3612064}
\usepackage{multirow}
\usepackage{tabularx}
\usepackage{graphicx}
\usepackage{svg}
\usepackage{algorithm}
\usepackage{algorithmic}
\usepackage{graphicx}
\usepackage{float}
\usepackage{subfig}

\usepackage{soul}
\usepackage{color}

\begin{document}

\title{Normality Learning-based Graph Anomaly Detection via Multi-Scale Contrastive Learning}

\author{Jingcan Duan}
\email{jingcan\_duan@163.com}
\affiliation{
  \institution{National University of Defense Technology}
  \city{Changsha}
  \state{Hunan}
  \country{China}}

\author{Pei Zhang}
\affiliation{
  \institution{National University of Defense Technology}
  \city{Changsha}
  \state{Hunan}
  \country{China}}

\author{Siwei Wang}
\authornotemark[1]
\affiliation{
  \institution{Intelligent Game and Decision Lab}
  \city{Beijing}
  \country{China}}

\author{Jingtao Hu}
\affiliation{
  \institution{National University of Defense Technology}
  \city{Changsha}
  \state{Hunan}
  \country{China}}

\author{Hu Jin}
\affiliation{
  \institution{National University of Defense Technology}
  \city{Changsha}
  \state{Hunan}
  \country{China}}

\author{Jiaxin Zhang}
\affiliation{%
  \institution{National University of Defense Technology}
  \city{Changsha}
  \state{Hunan}
  \country{China}}

\author{Haifang Zhou}
\authornotemark[1]
\affiliation{%
  \institution{National University of Defense Technology}
  \city{Changsha}
  \state{Hunan}
  \country{China}}

\author{Xinwang Liu}
\authornote{Corresponding authors.}
\affiliation{%
  \institution{National University of Defense Technology}
  \city{Changsha}
  \state{Hunan}
  \country{China}}

\renewcommand{\shortauthors}{Jingcan Duan et al.}

\begin{abstract}
Graph anomaly detection (GAD) has attracted increasing attention in machine learning and data mining. Recent works have mainly focused on how to capture richer information to improve the quality of node embeddings for GAD. Despite their significant advances in detection performance, there is still a relative dearth of research on the properties of the task. GAD aims to discern the anomalies that deviate from most nodes. However, the model is prone to learn the pattern of normal samples which make up the majority of samples. Meanwhile, anomalies can be easily detected when their behaviors differ from normality. Therefore, the performance can be further improved by enhancing the ability to learn the normal pattern. To this end, we propose a normality learning-based GAD framework via multi-scale contrastive learning networks (NLGAD for abbreviation). Specifically, we first initialize the model with the contrastive networks on different scales. To provide sufficient and reliable normal nodes for normality learning, we design an effective hybrid strategy for normality selection. Finally, the model is refined with the only input of reliable normal nodes and learns a more accurate estimate of normality so that anomalous nodes can be more easily distinguished. Eventually, extensive experiments on six benchmark graph datasets demonstrate the effectiveness of our normality learning-based scheme on GAD. Notably, the proposed algorithm improves the detection performance (up to 5.89\% AUC gain) compared with the state-of-the-art methods. The source code is released at \href{https://github.com/FelixDJC/NLGAD}{https://github.com/FelixDJC/NLGAD}.
\end{abstract}

\ccsdesc[500]{Computing methodologies~Anomaly detection}
\ccsdesc[500]{Mathematics of computing~Graph algorithms}

\keywords{Graph Anomaly Detection, Normality Learning, Multi-Scale Contrastive Learning}


\maketitle

\section{Introduction}
Recently, graph anomaly detection (GAD) has become an increasing application of graph-based machine learning for researchers~\cite{wu2020comprehensive,liu2022graph,ma2021comprehensive}. Well-established graph anomaly detection algorithms can effectively detect anomalous samples whose behavior obviously strays from the majority of nodes in a graph. Owing to its excellent performance in preventing real-world harmful situations, GAD has been used in a wide range of areas, including social network anomaly~\cite{heard2010bayesian,yu2016survey}, social spam detection~\cite{rao2021review}, medical domain~\cite{han2021madgan}, network intrusion detection~\cite{zhou2021hierarchical}, etc. Structured graph data in GAD contains both node feature information and network structure information. The mismatch between such two types of information generates two typical anomalies, i.e. contextual and structural anomalies~\cite{ding2019deep,liu2021anomaly,liu2022bond}. The former is the nodes that are dissimilar to their neighbors. The latter indicates that some nodes have dissimilar features but are unusually closely connected.

\begin{figure}[!t]
\centering
\subfloat[Cora-AT.]{
\includegraphics[width=0.24\textwidth]{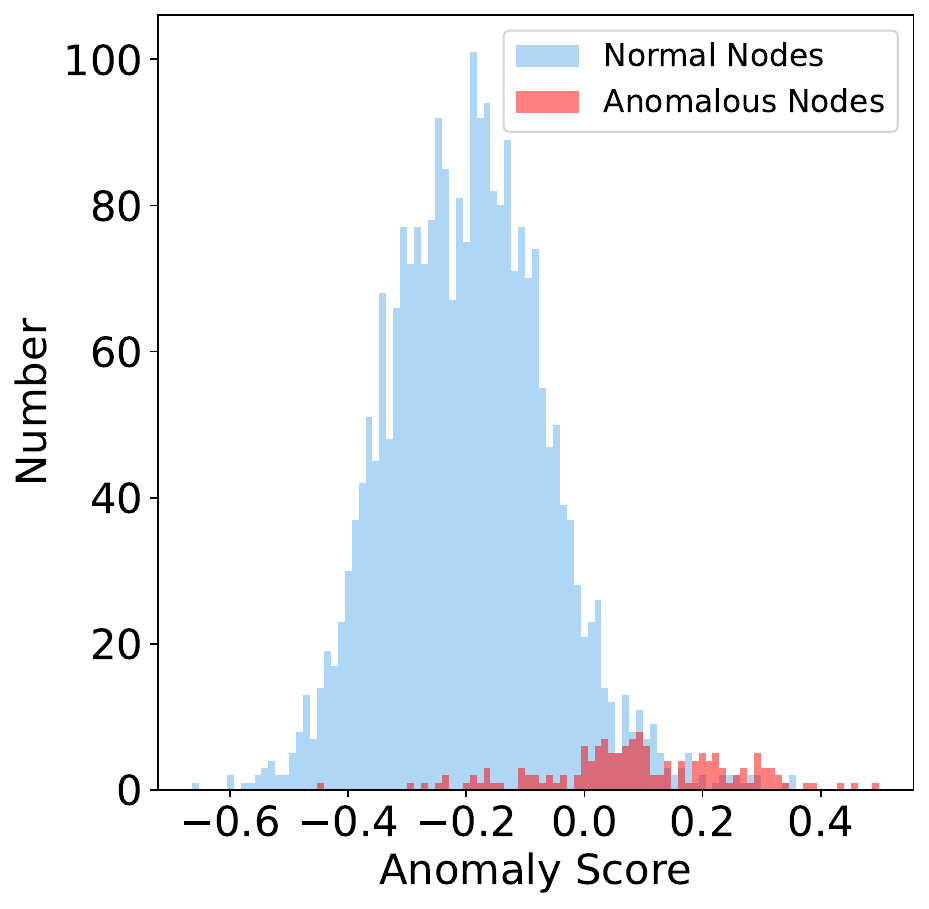}
\label{fig:distri:a}
}
\subfloat[Cora-NT.]{
\includegraphics[width=0.24\textwidth]{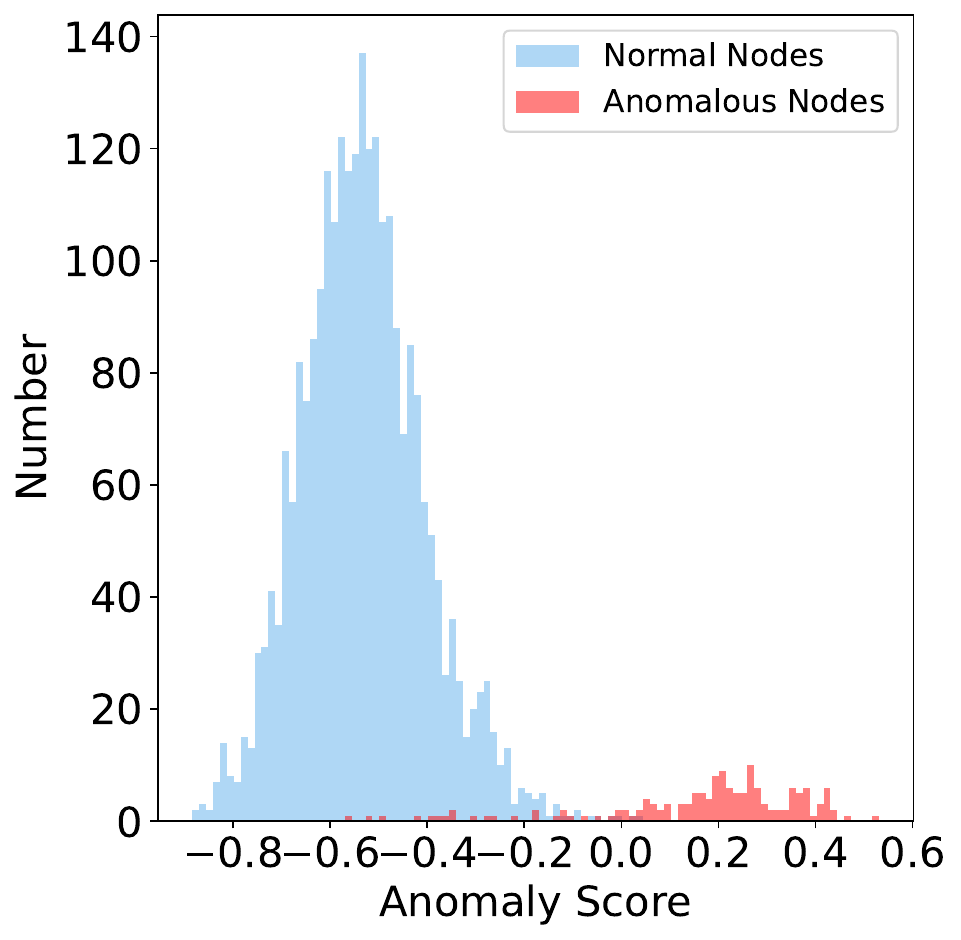}
\label{fig:distri:b}
}\\
\subfloat[DBLP-AT.]{
\includegraphics[width=0.24\textwidth]{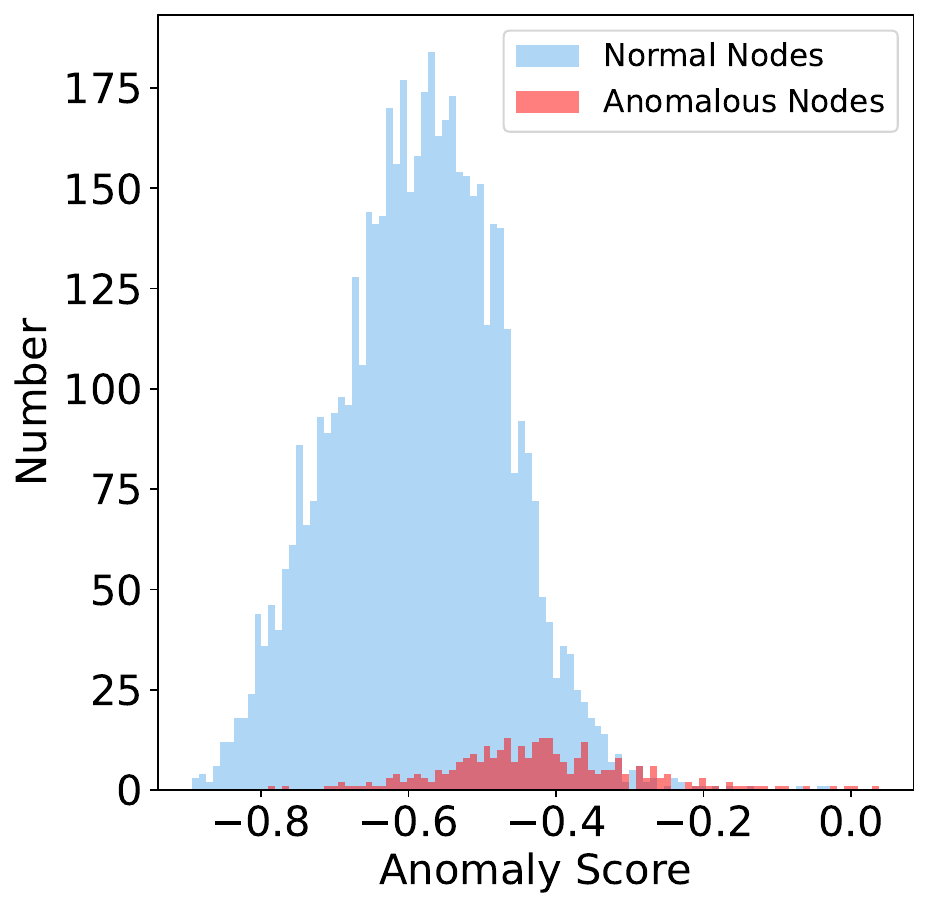}
\label{fig:distri:c}
}
\subfloat[DBLP-NT.]{
\includegraphics[width=0.24\textwidth]{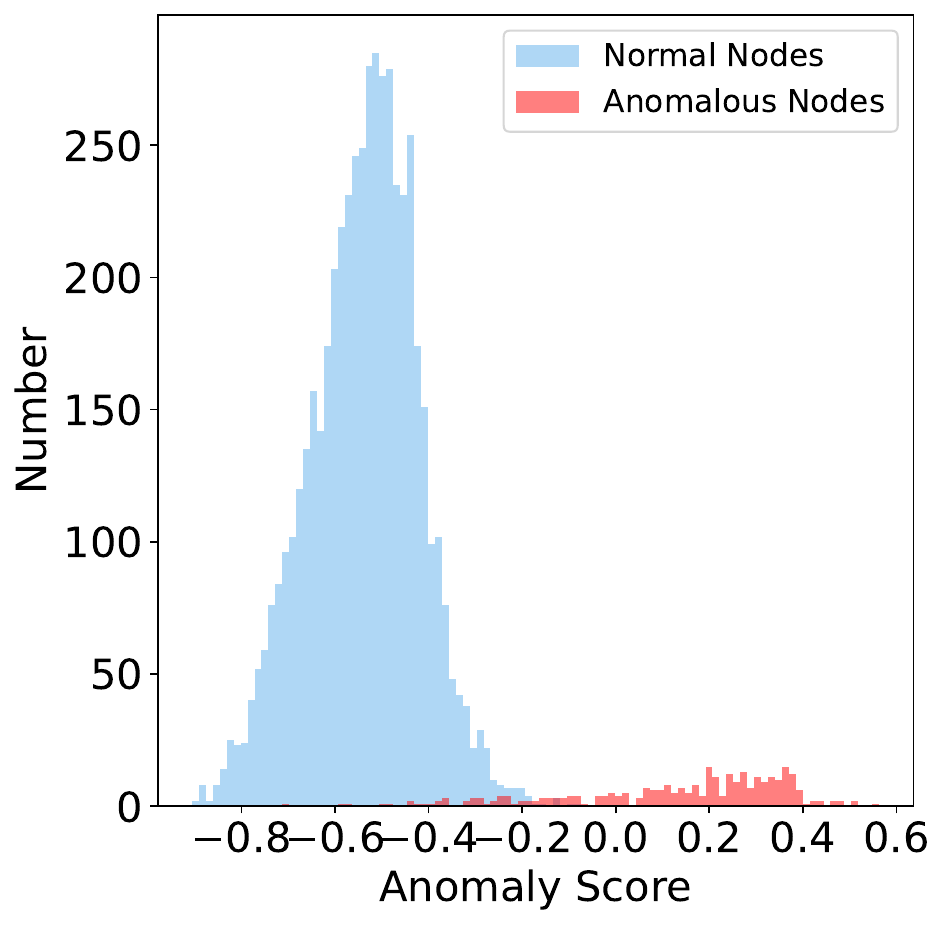}\label{fig:distri:d}
}
\caption{Anomaly score distribution diagrams. These subfigures are the distributions of anomaly scores computed by ANEMONE for nodes on Cora and DBLP. In the model training phase, (a) and (c) use all nodes as the training input (AT). Differently, (b) and (d) only input the normal nodes for training (NT).}
\label{fig:distribution}
\vspace{-15pt}
\end{figure}

Earlier researches improve anomaly detection performance through feature engineering, which cannot handle high-dimensional node features in graph. In recent years, the emergence of graph neural networks (e.g. GCN~\cite{kipf2016semi}) has brought hope for solving the appealing problem~\cite{liuyue_SCGC, liuyue_DCRN, liuyue_HSAN, liuyue_survey, liuyue_Dink_net}. Researchers instinctively introduce it to GAD. DOMINANT~\cite{ding2019deep} designs an ingenious GCN-based autoencoder. Then it compares the original feature and adjacency matrices with reconstructed matrices after GCN. Anomalous nodes have greater variations than normal ones. DOMINANT achieves a great comparative advantage over earlier methods. Nevertheless, GCN is a low-pass filter in essence~\cite{bo2021beyond} and will smooth the anomalies with their neighbors, which does harm to the GAD task. To overcome this inherent defect, researchers redesign specific graph filters which obtain anomalous information from different frequencies~\cite{tang2022rethinking,chaican}. As with the other semi-supervised approaches~\cite{liang2018semi,kumagai2021semi}, they still suffer from the expensive label-collection progress. Without the reliance on ground-truth labels of known anomalies, the pioneering work CoLA~\cite{liu2021anomaly} first introduces the graph contrastive learning pattern to GAD. It digs local feature and structure information through subgraph-node contrast. Normal nodes used to be similar to their neighborhoods in positive pairs and dissimilar in negative pairs, while anomalies behave differently. Subsequent works~\cite{jin2021anemone,zheng2021generative,zhang2022reconstruction} improve CoLA by adding node-node contrast, reconstruction loss, and global information, respectively. Overall, they focus on improving the quality of node representations through diverse types of information for GAD.

However, existing works ignore the characteristics of graph anomaly detection. As indicated in Figure~\ref{fig:distribution}, we draw the distribution diagrams of the anomaly scores computed by ANEMONE~\cite{jin2021anemone} on datasets Cora and DBLP. In Subfigure~\ref{fig:distri:a} and~\ref{fig:distri:c}, \textbf{AT} represents that all nodes are used as the training input, while \textbf{NT} means only the normal nodes are treated as the training input in Subfigure~\ref{fig:distri:b} and~\ref{fig:distri:d}. It is worth noting that the anomaly score is calculated from all nodes in the inference phase. The intuitive observation is that the distributions of normal nodes (blue) are further away from the distributions of anomalies (red) in NT than in AT. Usually, the smaller overlap region between these two distributions means that the trained model has a stronger anomaly detection capability. Since normal nodes make up the majority of samples in a graph, GAD models are more prone to learn the pattern of normal samples. It is summarized as \textbf{Normality Learning} in GAD. Nodes that deviate from this pattern are usually recognized as anomalous ones. It is a reasonable approach to boost detection performance by further enhancing the ability to learn the normal pattern. The phenomenon in Figure~\ref{fig:distribution} demonstrates that if only normal nodes are utilized in the training phase, the ability of the model to learn the normal pattern will be tremendously enhanced. Conversely, mixing anomalous and normal nodes will harm the normality learning and weaken the detection performance. Based on these analyses, how to improve detection performance utilizing normality learning under an unsupervised paradigm in GAD is still an issue worth investigating.

\begin{figure*}[htbp]
    \centering
    \includegraphics[width = \textwidth]{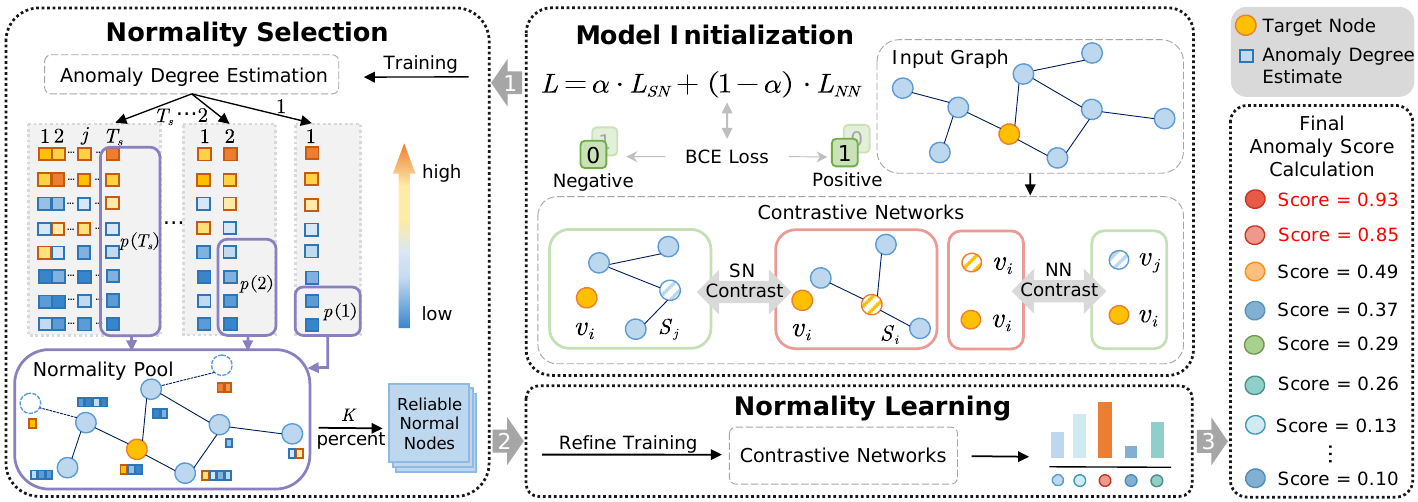}
    \caption{The overview framework of NLGAD.     It consists of three modules: (1) Model Initialization: We initialize the model by multi-scale contrastive networks which have subgraph-node (SN) and node-node (NN) contrasts; (2) Normality Selection: We train the model with all nodes and perform anomaly degree estimation at each step. Then, we perform a hybrid strategy and assign the normal pseudo-labels to reliable nodes; (3) Normality Learning: We input the reliable normal nodes into the model and re-train it. Finally, the refined model is used to calculate the final anomaly score for each node.}
    \label{fig:model}
\vspace{-10pt}
\end{figure*}

To this end, we propose a \textbf{N}ormality \textbf{L}earning-based \textbf{G}raph \textbf{A}nomaly \textbf{D}etection framework via multi-scale contrastive learning networks termed \textbf{NLGAD}. The entire framework is shown in Figure~\ref{fig:model}. To initialize the model, we first construct the contrastive networks containing subgraph-node (SN) and node-node (NN) contrasts, which can fuse multi-scale of anomalous information in graph. The model is trained using a two-phase scheme. In the normality selection phase, we leverage all nodes as the input to train the model. In this process, we devise a hybrid selection strategy that assigns normal pseudo-labels to sufficient nodes. To be specific, we evaluate the anomaly degree for each node at each step and then add the high-confidence estimates to the normality pool with the improvement of model capability. At the end of this phase, we synthetically compute the anomaly estimate for each node and assign normal pseudo-labels to the lowest part of the nodes. In the normality learning phase, the model is retrained with the input of selected normal nodes based on the normality pool. By taking advantage of normal learning, the network will be further refined for the task. Finally, we calculate the final anomaly scores for all nodes. Furthermore, it is remarkable that NLGAD does not require supervision information from the ground-truth labels throughout the training phase. In summary, our contributions can be summarized in the followings:

\vspace{-10pt}
\begin{itemize}
\item We empirically investigate the different impacts of AT and NT on GAD and explain the superiority of normality learning in this task.
\item We propose a normality learning-based graph anomaly detection scheme on attributed networks, without any manual annotation.
\item A novel hybrid normality selection strategy is designed to pick out sufficient and reliable normal nodes to provide support for normality learning.
\item Experiments demonstrate the notable advantage of NLGAD against current graph anomaly detection competitors, which indicates the effectiveness of normality learning. And the ablation studies further validate the necessity of the normality selection strategy. 
\end{itemize}

\section{Related work}
In the following section, we provide a brief review of graph anomaly detection via deep learning. Graph neural networks (GNNs)~\cite{kipf2016semi, peng2021attention, liu2021self, wang2022gcl} have significant advantages in capturing deeper information of graph datasets. Based on GNNs, \cite{zhou2021subtractive,wang2021one} extend the traditional anomaly detection algorithm one-class SVM to the graph datasets. Graph generative models have also achieved notable improvement. DOMINANT~\cite{ding2019deep} first adopts a GCN-based autoencoder in graph anomaly detection and treats the nodes with larger reconstruction errors as anomalies. Based on DOMINANT, researchers boost the detection performance by fusing the community information of each node~\cite{luo2022comga}. \cite{tang2022rethinking,chaican} enhance the capability of high-pass graph filters and relieve the over-smoothing phenomenon carried by GCN. Self-supervised learning is an important branch of unsupervised learning. HCM~\cite{huang2021hop} regards the predictive hops of the target node and its neighborhoods as its anomalous degree. The larger the number of predictive hops, the higher probability of being anomalies. The pioneer~\cite{liu2021anomaly} leverages the graph contrastive learning pattern for the first time and captures anomalies from node-subgraph comparison. It has become a strong baseline in graph anomaly detection. Based on it, ANEMONE~\cite{jin2021anemone} proposes a new node-node contrastive pattern. GRADATE~\cite{duan2023graph} optimizes the subgraph embeddings in GAD by adding a subgraph-subgraph contrast. Differently, SL-GAD~\cite{zheng2021generative} combines the advantages of reconstruction and graph contrastive learning paradigms. Sub-CR~\cite{zhang2022reconstruction} captures more anomalous information by utilizing a newly generated global view. \cite{jingcan2022gadmsl} focuses more on detecting structural anomalies.

\section{Problem Definition}
In this section, we formally define the main problem. The main notations are summarized in Table~\ref{table:symbol}.

\textit{Problem definition (Graph Anomaly Detection).} An undirected attributed graph $\mathcal{G} = \left ( \mathcal{V}, \mathcal{E} \right ) $ is composed of: (1) a set of nodes $\mathcal{V}$, where $\left |\mathcal{V}\right | = n$; (2) a set of edges $\mathcal{E}$, where $\left |\mathcal{E}\right | = m$. Specifically, the graph is formalized as its feature matrix $X$ and adjacency matrix $A$. In addition, the adjacency matrix $\mathbf{A}_{ij} = 1$ represents there is an edge between node $v_{i}$ and $v_{j}$; otherwise, $\mathbf{A}_{ij} = 0$. In graph anomaly detection, the model is trained to learn a specific function $f\left ( \cdot  \right ) $ that can calculate the anomaly score $score\left ( v_{i} \right )$ for the node $v_{i}$. The larger the anomaly score, the more likely it is that the node is anomalous.

\begin{table}[ht]
\centering
\caption{Table of main symbols.}
\begin{tabularx}{0.50\textwidth}{p{2.0cm}<{\centering}p{6.0cm}}
\toprule
$\textbf{Notations}$ & $\textbf{Definitions}$\\
\midrule
$\mathcal{G}$       & An undirected attributed graph  \\
$v_{i}$       & The $i$-th node of $\mathcal{G}$\\
$\mathbf{A}\in\mathbb{R}^{n\times n}$       & Adjacency matrix of $\mathcal{G}$\\
$\mathbf{D}\in\mathbb{R}^{n\times n}$&Degree matrix of $\mathbf{A}$\\
$\mathbf{X}\in\mathbb{R}^{n\times d}$       & Feature matrix of $\mathcal{G}$\\
$\boldsymbol{x}_{i}\in\mathbb{R}^{1\times d}$     & Feature vector of $v_{i}$ that $\boldsymbol{x_{i}}\in\mathbf{X}$\\
$\mathbf{H}^{\left ( \ell \right ) }\in\mathbb{R}^{n\times d^{\prime}}$&Subgraph hidden representation of the ${\ell}$-th layer\\
$\boldsymbol{h}^{\left ( \ell \right ) }\in\mathbb{R}^{1\times d^{\prime}}$&Node hidden representation of the ${\ell}$-th layer\\
$\mathbf{W}^{\left ( \ell \right )}\in\mathbb{R}^{d^{\prime}\times d^{\prime}}$&Network parameters of the ${\ell}$-th layer\\
$y_{i}\in \left \{ 0,1 \right \} $&Label of instance pair in BCE loss\\
$score\left(v_{i}\right)$&Final anomaly score of $v_{i}$\\
\bottomrule
\end{tabularx}
\label{table:symbol}
\vspace{-10pt}
\end{table}

\section{Method}
In the following section, we present the proposed framework, NLGAD. We first sample subgraphs around nodes and form subgraph-node contrast. The model also leverages node-node contrast to capture the node-level anomalous information. After that, the model will be initialized. Then, we train the model through two phases. During the normality selection phase, the model is trained with all nodes and estimates the anomaly degree for each node. We add high-confident estimates into the normality pool by a tailored dynamic strategy and retain the lowest part of nodes. During the second phase, we refine the model with the selected normal nodes and then calculate the final anomaly scores for all nodes. Finally, we perform complexity analysis on NLGAD.

\subsection{Model Initialization}
The multi-scale graph contrastive pattern has been proven valid for GAD~\cite{jin2021anemone}. Thus, we build a graph contrastive network under the multi-scale strategy as the backbone of our model. To be specific, we first adopt random walk to sample subgraphs around nodes. Then, we construct the subgraph-node and node-node contrasts separately. The former contrast captures anomalous information from node neighborhoods. The second one further digs out the node-level anomalies. Based on both contrasts, the model will be initialized by information from nodes and their neighborhoods.

\subsubsection{Subgraph-Node (SN) contrast.}
Firstly, we sample the subgraphs around the nodes to create subgraph-node contrasts. The feature and structure information from the subgraphs can represent the local neighborhood information of the nodes, which is useful for discerning anomalies. In subgraph-node contrast, the target node $v_{i}$ forms positive pairs with the subgraphs where it is located and forms negative pairs with the subgraphs where another node is located. Inspired by~\cite{tong2006fast,perozzi2014deepwalk}, we adopt random walk with restart (RWR) as the subgraph sampling method.

We utilize a GCN layer to learn the subgraph representations in the latent space. It is worth noting that the features of the target node $v_{i}$ in the subgraph have been masked to 0, which is to avoid the smoothing of anomalous information. The subgraph representations in the hidden layer can be denoted as:

\begin{equation}
\mathbf{H}^{\left (\ell+1  \right ) }_{i} =\sigma \left ( \widetilde{\mathbf{D}}^{-\frac{1}{2}}_{i}\widetilde{\mathbf{A}}_{i}\widetilde{\mathbf{D}}^{-\frac{1}{2}}_{i}\mathbf{H}^{\left (\ell  \right ) }_{i}\mathbf{W}^{\left ( \ell \right ) }\right ),
\label{subgraph_GCN}
\end{equation}
where $\mathbf{H}^{\left (\ell+1  \right ) }_{i}$ and $\mathbf{H}^{\left (\ell  \right ) }_{i}$ are the hidden representation of the ${\left (\ell+1\right ) }$-th and ${\ell}$-th layer, $\widetilde{\mathbf{A}}_{i}$ is the subgraph adjacency matrix with self-loop, $\widetilde{\mathbf{D}}_{i}$ is the degree matrix of $\widetilde{\mathbf{A}}_{i}$, $\mathbf{W}^{\left (\ell \right ) }$ indicates the network parameters. And $\sigma \left (\cdot   \right )$ is the activation function ReLU here.

We adopt a \textit{Readout} function to obtain the final subgraph representation $\boldsymbol{z}_{i}$. In practice, an average function is used to achieve it. The final representation is defined as:
\begin{equation}
\boldsymbol{z}_{i}=Readout\left (\mathbf{Z}_{i}\right ) =\frac{1}{n_{i}}\sum_{j=1}^{n_{i}} \left (\mathbf{Z}_{i} \right )_{j}.
\end{equation}

In the meantime, we utilize a MLP to map the target node features to the same latent space as the subgraph:
\begin{equation}
\boldsymbol{h}^{\left (\ell+1  \right ) }_{i} =\sigma \left (\boldsymbol{h}^{\left (\ell  \right ) }_{i}\mathbf{W}^{\left (\ell \right ) }\right ),
\end{equation}
where $\mathbf{W}_{i}^{\left (\ell \right ) }$ is shared with GCN in Eq.~\eqref{subgraph_GCN}. The final embedding of the target node $v_{i}$ is denoted as $\boldsymbol{e}_{i}$.

The similarity $s_{i}$ of the target node $v_{i}$ and its subgraph is directly related to its anomaly degrees~\cite{liu2021anomaly}. We use a \textit{Bilinear} function to measure it:
\begin{equation}
s_{i}  =Bilinear\left (\boldsymbol{z}_{i},\boldsymbol{e}_{i}\right ).
\label{node_subgraph_similarity}
\end{equation}

\subsubsection{Node-Node (NN) contrast.}



To get diverse information, we leverage the node-node contrast to capture node-level anomalies. The target node embedding aggregated from the other nodes in the subgraph forms a positive pair with the same node after MLP, and forms a negative pair with another node after MLP. Same with the subgraph-node contrast, the target node features in the subgraph will be masked. We adopt a new GCN to obtain the subgraph's representation $\mathbf{H}^{\prime \left (\ell \right ) }_{i}$, while $\boldsymbol{h}^{\prime \left (\ell \right ) }_{i} = \mathbf{H}^{\prime \left (\ell \right ) }_{i}\left [ 1,: \right ] $ is the hidden representation of $v_{i}$. $\boldsymbol{u}_{i}$ is the final embedding of target node. Similarly, we utilize a new MLP to map the node features into the same space. And $\hat{\boldsymbol{e}}_{i}$ is the target node's final embedding. Then we can measure the similarity of $\boldsymbol{u}_{i}$ and $\hat{\boldsymbol{e}}_{i}$ through a \textit{Bilinear} function.

\subsubsection{Loss function.}
In subgraph-node contrast, the target node tends to be similar to the subgraph in the positive pair, i.e. $s_{i} =1$. Differently, it tends to be dissimilar to the subgraph in the negative pair, i.e. $s_{i} =0$. It is natural to use the binary cross-entropy (BCE) loss to train the networks:
\begin{equation}
\mathcal{L}_{SN}=- \sum_{i=1}^{n}\left (y_{i}\log{\left (s_{i} \right )} +  \left (1 - y_{i}\right )\log{\left ( 1 - s_{i} \right ) }\right ),
\label{node_subgraph_loss1}
\end{equation}
where $y_{i}$ is equal to 1 in positive pairs and is equal to 0 in negative pairs. The node-node contrast loss function $\mathcal{L}_{NN}$ can also be defined like $\mathcal{L}_{NN}$.


After that, we fuse the diverse anomalous information with a comprehensive loss function:

\begin{equation}
\mathcal{L}= \alpha \cdot \mathcal{L}_{SN}+\left ( 1-\alpha  \right ) \cdot \mathcal{L}_{NN},
\label{total_loss}
\end{equation}
where ${\alpha \in \left ( 0,1 \right ) }$ is a trade-off parameter to balance the importance between two contrasts.

\subsection{Normality Selection}
Previous unsupervised approaches in other fields~\cite{gupta2019unsupervised,park2021improving,zhan2021mutual} retain only a small number of nodes with high-confident pseudo-labels. However, normality learning in GAD requires a sufficient number of normal nodes to train the model and learn a more accurate estimate for the normal pattern. Therefore, the normality selection strategy should consider not only the reliability but also the number of normal nodes. In this phase, we propose a new hybrid strategy to conduct normality selection by inputting all the nodes into the training model. Specifically, we first employ a \textbf{dynamic strategy} that conducts \textbf{anomaly degree estimation} for each node at each step and gradually adds high-confident estimates to the normality pool. A \textbf{percent strategy} is then used to assign normal pseudo-labels to the lowest portion of nodes.

\subsubsection{Anomaly degree estimation.} 
At each step, we estimate a rough anomaly degree for each node. In subgraph-node contrast, a normal node tends to be similar to the subgraph in the positive pair and dissimilar to the one in the negative pair. Conversely, an anomalous node is dissimilar from the subgraphs in both positive and negative pairs. Therefore, the anomaly degree of the target node can be represented as:

\begin{equation}
E_{SN}\left ( v_{i} \right ) = s_{i}^{n} - s_{i}^{p},
\label{anomaly_score1}
\end{equation}
where $s_{i}^{p}$ and $s_{i}^{n}$ represent the similarity of positive and negative pairs, respectively.

Likewise, we can also compute the anomaly degree estimate $E_{NN}\left ( v_{i} \right )$ in node-node contrast. By integrating the anomalous information from two contrasts, we calculate the anomaly degree estimate $E\left ( v_{i} \right ) $ for the target node $v_{i}$:
\begin{equation}
E\left ( v_{i} \right ) =\alpha\cdot E_{SN}\left ( v_{i} \right )  +\left ( 1-\alpha  \right ) \cdot E_{NN}\left ( v_{i} \right ),
\label{anomaly_score2}
\end{equation}
where $\alpha$ is shared with Eq.~\eqref{total_loss}.

\subsubsection{Dynamic strategy.}
The detection capability of the model will increase as the network is trained. Therefore, the anomaly degree estimates calculated at each step are not fully reliable. We adopt a dynamic strategy to retain the most reliable part of the results. As shown in Subfigure.~\ref{fig:distri:a} and~\ref{fig:distri:c}, lots of normal nodes are usually erroneously assigned higher anomaly scores by GAD models, which confuses them with anomalous nodes. Conversely, the lowest part of the anomaly scores is rarely misclassified as anomalous nodes. This suggests that lower values in the distribution of anomaly scores have a higher confidence level. Therefore, we gradually store the nodes with the lowest proportion of anomaly degree estimates in the normality pool. We utilize a speed function $p\left ( j \right ) $ to specify the number of stored estimates at step $j$. For the characteristics of the task, we argue that this vital function must satisfy the following conditions: \textbf{(1) $p\left ( j \right ) $ starts with a smaller value.} It corresponds to the initial low detection performance of the model. A start value that is too large will add more unreliable estimates to the normality pool. \textbf{(2) $p\left ( j \right ) $ is monotonically and slowly increasing with respect to $j$.} Detection performance increases slowly as the model is trained. It is natural to make $p\left ( j \right ) $ continue to grow. Meanwhile, a small growth rate can avoid adding marginal anomaly estimates, ensuring that the normality pool retains reliable estimates. \textbf{(3)$p\left ( j \right ) $ ends with a larger value.} This works well with the percent strategy and could assign normal pseudo-labels to as many reliable nodes as possible. In practice, we adopt a function from the $\tan \left ( \cdot \right )$ family to achieve it:

\begin{equation}
p\left ( j \right ) = n \cdot \tan \left ( \frac{\pi}{4} \cdot \frac{j}{T_{s}} \right ).
\label{polynomial}
\end{equation}
The anomaly degree estimates of all nodes at each step are normalized to $\left [ 0, 1 \right ] $. $p\left ( j \right )$ represents the number of node anomaly estimates added to the normality pool at step $j$, $n$ is the number of nodes in the graph, and $T_{s}$ is the number of training steps in the normality selection module. It is worth noting that the other functions can also be applied to $p\left ( j \right )$. We leave this technical extension for future work.

\subsubsection{Percent strategy.}
A single strategy cannot assign high-confident pseudo-labels to nodes. We design a percent strategy to label nodes while leaving enough nodes as normal nodes. To mitigate the randomness of different steps and improve the reliability of the anomaly degree for each node, we average their multi-step anomaly estimates in the normality pool:

\begin{equation}
E\left(v_{i}\right)^{\prime}=\frac{1}{m_{i}}\sum_{q=1}^{m_{i}}E\left ( v_{i} \right )_{q},
\label{temp_score1}
\end{equation}
where $m_{i}$ is the times of $v_{i}$'s anomaly degree estimates added to the normality pool. After that, we sort the estimates in ascending order and return the index vector, 
\begin{equation}
\boldsymbol{b} = arg\text{ }sort\left \{ E\left(v_{1}\right)^{\prime},E\left(v_{2}\right)^{\prime},\dots,E\left(v_{n}\right)^{\prime}\right \},
\label{temp_score2}
\end{equation}
where $\boldsymbol{b}$ is the index vector. Then, the nodes with the lowest $K$ percent estimates will be assigned normal pseudo-labels.

\subsection{Normality Learning}
\subsubsection{Refine training.}
After obtaining reliable normal nodes, we refine the model according to a normality learning-based scheme. Its purpose is to boost the model's ability to learn the pattern of normal nodes. Then we can obtain a more accurate estimate of normality. After that, in the inference phase, the anomaly score distribution of normality will be further away from the distribution of anomalies. In this phase, we utilize the nodes with normal pseudo-labels to retrain the model with $T_{r}$ epochs.

\subsubsection{Final anomaly score calculation.} Finally, in the inference phase, we use the model to compute the final anomaly score for each node. One detection with RWR will miss much semantic information. Hence, we employ a multi-round detection strategy. In each round, we calculate the $score\left(v_{i}\right)$ for $v_{i}$ via Eq.~\eqref{anomaly_score1}\eqref{anomaly_score2}. Inspired by~\cite{jin2021anemone}, we compute the final anomaly score:

\begin{equation}
\begin{split}
score\left(v_{i}\right)_{mean}&=\frac{1}{r}\sum_{k=1}^{r}score\left(v_{i}\right)^{\left (k \right )}, \\
score\left(v_{i}\right)_{std}&=\sqrt{\frac{1}{r}\sum_{k=1}^{r}\left (score\left(v_{i}\right)^{\left (k \right )}-score\left(v_ {i}\right)_{mean}\right )^{2}  },\\
score\left(v_{i}\right)&=score\left(v_{i}\right)_{mean}+score\left(v_{i}\right)_{std},
\end{split}
\label{anomaly_score3}
\end{equation}
where $r$ is the number of anomaly detection round, and $score\left(v_{i}\right)$ is the final anomaly score for $v_{i}$.

\subsection{Complexity Analysis}
\label{complexity}
We analyze the time complexity of each component in our method. The time complexity of each RWR subgraph sampling for all nodes is $\mathcal{O}\left ( c\delta n \right )$, where $c$ is the number of nodes within the subgraphs, $\delta$ is the mean degree of the network, and $n$ is the number of nodes in the graph. The time complexity of the GCN network for all nodes is $\mathcal{O}\left ( \left (Kqd+Kcd^{2}\right )n \right )$, where $K$ is the number of layers, $d$ is the dimension of node attributes in hidden space, and $q$ is the number of edges in subgraphs. For the training and inference phase, the overall \textbf{time complexity} of the proposed model is $\mathcal{O}\left (  n\left (c\delta+Kqd+Kcd^{2}\right )\left (T_{s}+T_{r}+ r\right ) \right )$, where $T_{s}+T_{r}$ is the total number of epochs in the training phase, $r$ is the round of detections in the inference phase.


\section{Experiment}
In this section, we perform extensive experiments on six widely-used benchmark datasets to verify the effectiveness of NLGAD. Firstly, we introduce the details of experimental settings and results. Then, we show the ablation studies of the hybrid strategy for normality selection and normality learning. Finally, we conduct sensibility analyses of hyper-parameters in the framework.

\subsection{Experimental Settings}
\subsubsection{Datasets} Table~\ref{table:datasets} lists six commonly used benchmark datasets in GAD. The datasets include Cora~\cite{sen2008collective}, UAI2010~\cite{wang2018unified}, CiteSeer~\cite{sen2008collective}, DBLP, Citation, and ACM~\cite{yuan2021higher}. UAI2010 is a graph dataset for community detection. The others are citation datasets, which contain citing relations of publications or coauthor relationships of researchers.

\begin{table}[ht]
\centering
\caption{The statistics of datasets.}
\begin{tabular}{ccccc}
\toprule
$\textbf{Datasets}$&$\textbf{Nodes}$&$\textbf{Edges}$&$\textbf{Features}$&$\textbf{Anomalies}$\\
\midrule
$\textbf{Cora}$& 2708 & 5429 & 1433 & 150 \\
$\textbf{UAI2010}$& 3067 & 28311 & 4973 & 150 \\
$\textbf{CiteSeer}$& 3327 & 4732 & 3703 & 150 \\
$\textbf{DBLP}$& 5484 & 8117 & 6775 & 300 \\
$\textbf{Citation}$& 8935 & 15098 & 6775 & 450 \\
$\textbf{ACM}$& 9360 & 15556 & 6775 & 450 \\
\bottomrule
\end{tabular}
\label{table:datasets}
\vspace{-10pt}
\end{table}

\subsubsection{Anomaly Injection.}
Our method aims at detecting the two typical types of anomalies in the graph. To verify its effectiveness, we follow~\cite{ding2019deep} to inject such two anomalies into the original graph:

\begin{itemize}
\item \textbf{Contextual anomalies} are injected by perturbing the feature values. In practice, we randomly select a node $v_{i}$ and another $n^{\prime}$ (fixed to 50) nodes. And we exchange $v_{i}$'s feature values with the most dissimilar node out of the $n^{\prime}$ nodes.
\item \textbf{Structural anomalies} are generated by perturbing the topological structure of the graph. To be specific, we randomly select $m^{\prime}$ nodes and make them fully connected. The number $m^{\prime}$ is usually 15.
\end{itemize}

We repeat the above two operations, injecting the same number of contextual and structural anomalies into the original graphs. The total number of anomalies in each dataset is shown in the last column of Table~\ref{table:datasets}.

\subsubsection{Metric.}
In the experiments, we employ the widely-used method \textbf{AUC} as the metric to evaluate the model. AUC is the under-line area of the ROC curve, which can represent the probability of selecting anomalies rather than normality in top anomaly score samples. The larger the AUC value, the better the model performance.

\subsubsection{Baselines.}
In this part, we compare NLGAD with eight well-known GAD methods, including LOF~\cite{breunig2000lof}, ANOMALOUS~\cite{peng2018anomalous}, DOMINANT~\cite{ding2019deep}, CoLA~\cite{liu2021anomaly}, ANEMONE~\cite{jin2021anemone}, SL-GAD~\cite{zheng2021generative}, HCM~\cite{huang2021hop}, and Sub-CR~\cite{zhang2022reconstruction}. Following CoLA, the used datasets in ANOMALOUS are reduced to 30 by PCA. It is worth noting that the first two methods are traditional shallow methods and the others work on deep neural networks.

\subsubsection{Model Settings.}
To balance performance and complexity, we set the size of subgraphs to 4. Both the features of nodes and subgraphs are mapped to 64 dimensions. The learning rate of the model is fixed at 0.001. We perform 700 epochs of total training on Cora, UAI2010, and CiteSeer, 900 epochs on DBLP, Citation, and ACM.

\begin{table*}[t]
\setlength{\tabcolsep}{5.5mm}
\caption{Performance comparison for AUC. The bold and underlined values indicate the best and runner-up results, respectively.}
\resizebox{\textwidth}{!}{
\begin{tabular}{c|c|cccccc}
\toprule
\makebox[0.1\textwidth][c]{\textbf{Category}}                & \textbf{Methods}          & \textbf{Cora}   & \textbf{UAI2010} & \textbf{CiteSeer} & \textbf{DBLP}   & \textbf{Citation} & \textbf{ACM}    \\ \midrule
\multirow{2}{*}{\textbf{Shallow}} & LOF (2000) & 0.3538 & 0.7052 & 0.3484 & 0.2694 & 0.3059 & 0.2843 \\
                                  & ANOMALOUS (2018) & 0.6688 & 0.7144 & 0.6581 & 0.6728 & 0.6356 & 0.5894 \\ \midrule
\multirow{7}{*}{\textbf{Deep}}    & DOMINANT (2019) & 0.8929 & 0.7698 & 0.8718 & 0.8034 & 0.7748 & 0.8152 \\
                                  & CoLA (2021) & 0.9065 & 0.7949 & 0.8863 & 0.7824 & 0.7296 & 0.8127 \\
                                  & ANEMONE (2021) & 0.9122 & \underline{0.8731} & 0.9227 & 0.8322 & 0.8028 & 0.8300 \\
                                  & SL-GAD (2021) & \underline{0.9192} & 0.8454 & 0.9177 & \underline{0.8461} & \underline{0.8095} & \underline{0.8450} \\
                                  & HCM (2021) & 0.6276 & 0.5210 & 0.6502 & 0.5572 & 0.5414 & 0.5507 \\
                                  & Sub-CR (2022) & 0.9133 & 0.8571 & \underline{0.9248} & 0.8061 & 0.7903 & 0.8428 \\ 
                                  & \textbf{NLGAD (Proposed)} & \textbf{0.9286} & \textbf{0.9320}  & \textbf{0.9457}   & \textbf{0.8524} & \textbf{0.8435}   & \textbf{0.8655} \\ \bottomrule

\end{tabular}}
\label{table:AUC}
\vspace{-5pt}
\end{table*}

\begin{figure*}[!ht]
\centering
\subfloat[Cora]{
\includegraphics[width=0.31\textwidth]{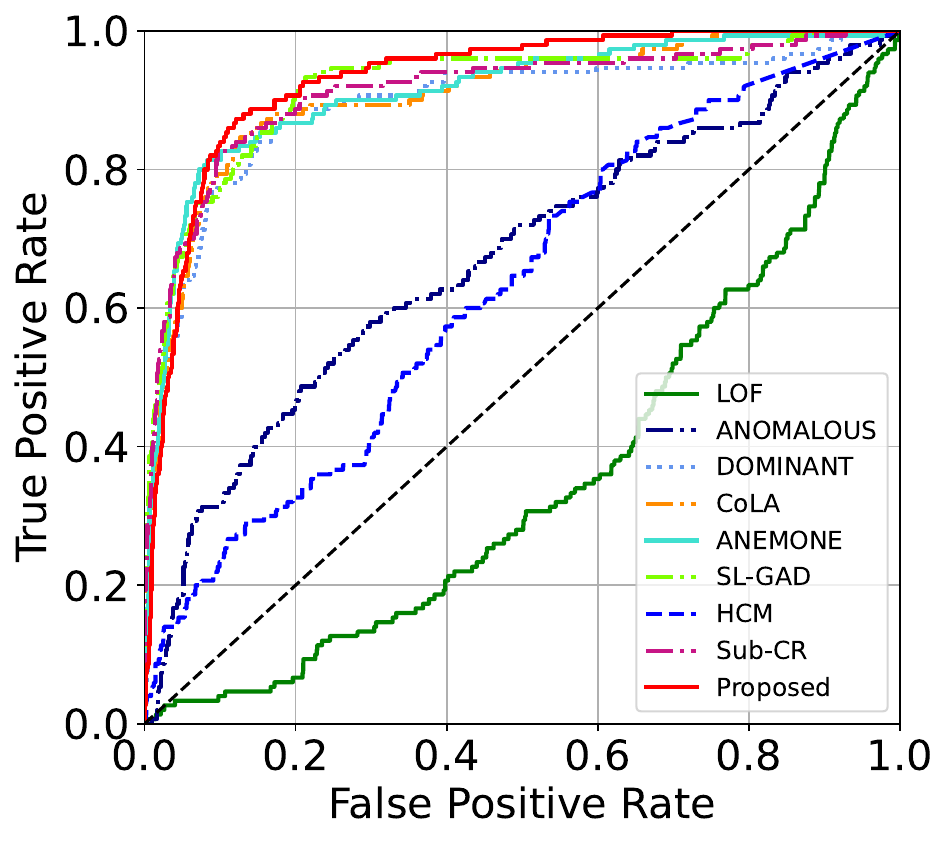}
}
\subfloat[UAI2010]{
\includegraphics[width=0.31\textwidth]{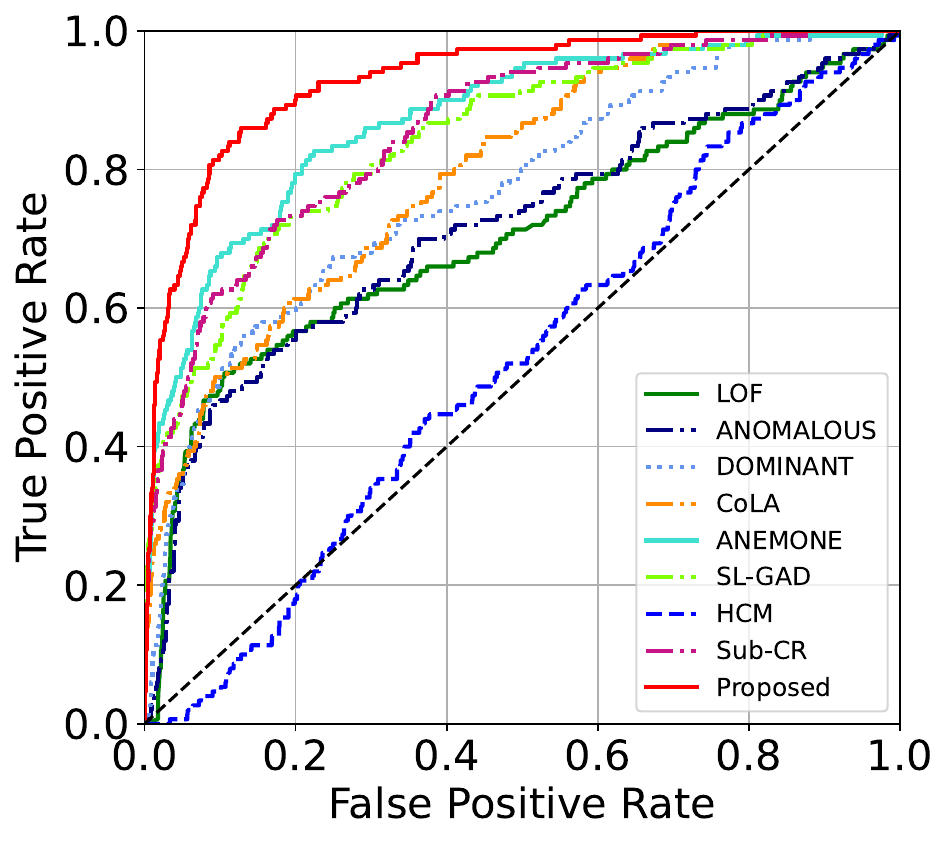}
}
\subfloat[CiteSeer]{
\includegraphics[width=0.31\textwidth]{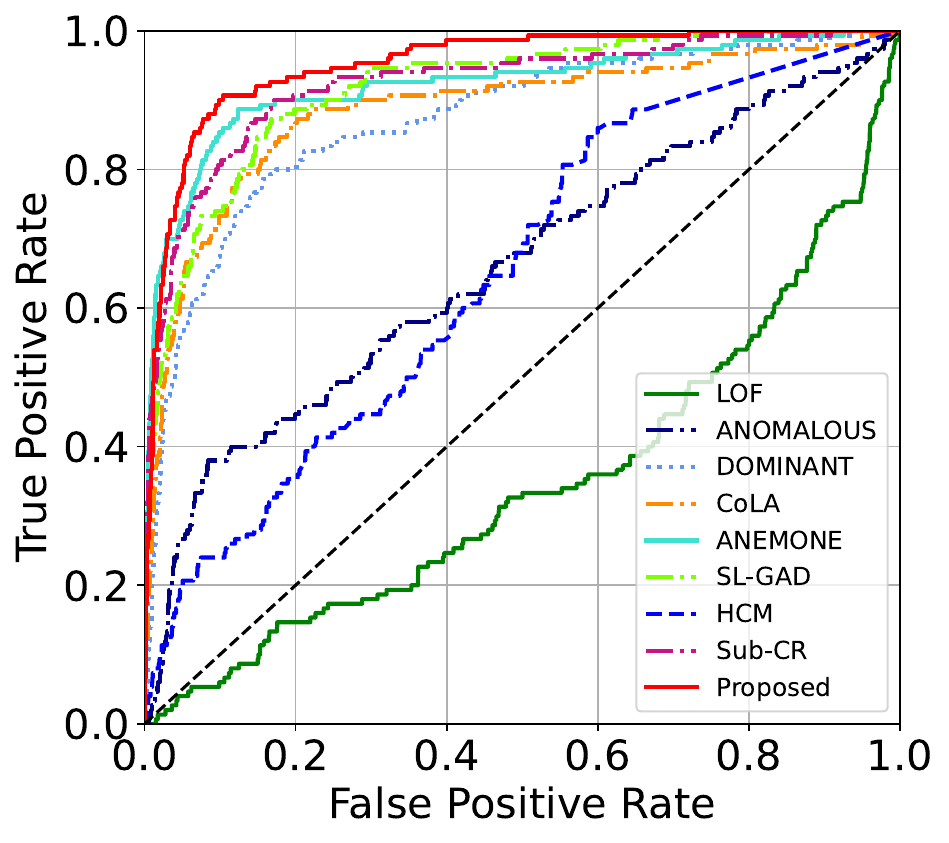}
}\\
\subfloat[DBLP]{
\includegraphics[width=0.31\textwidth]{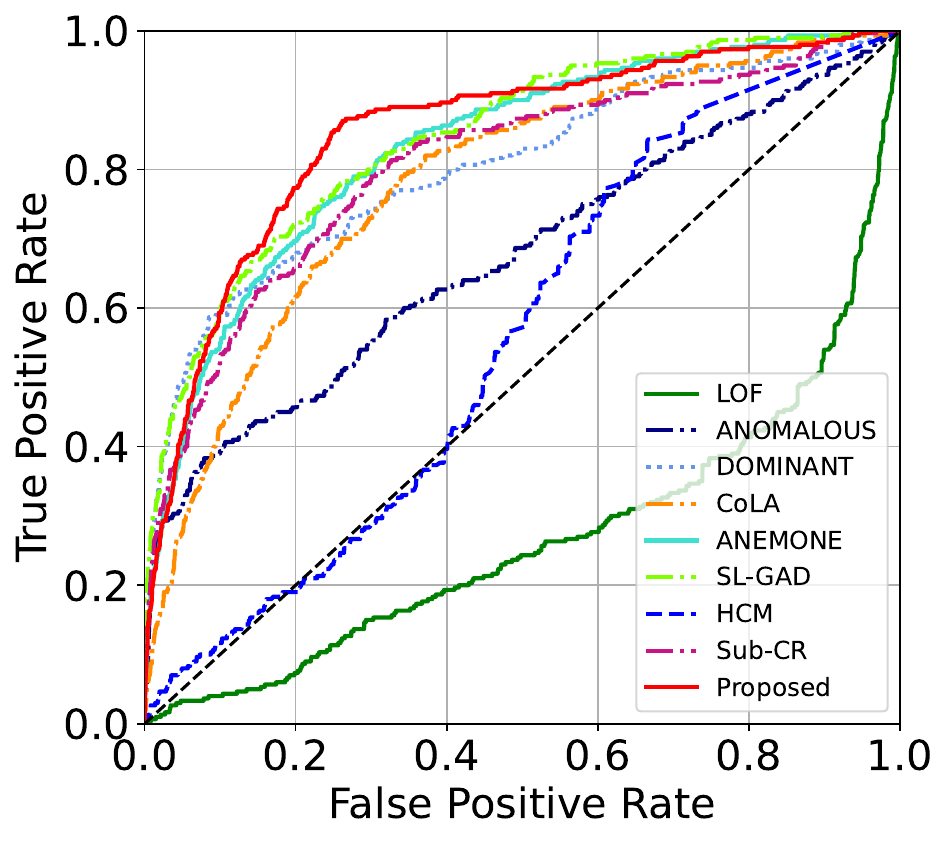}
}
\subfloat[Citation]{
\includegraphics[width=0.31\textwidth]{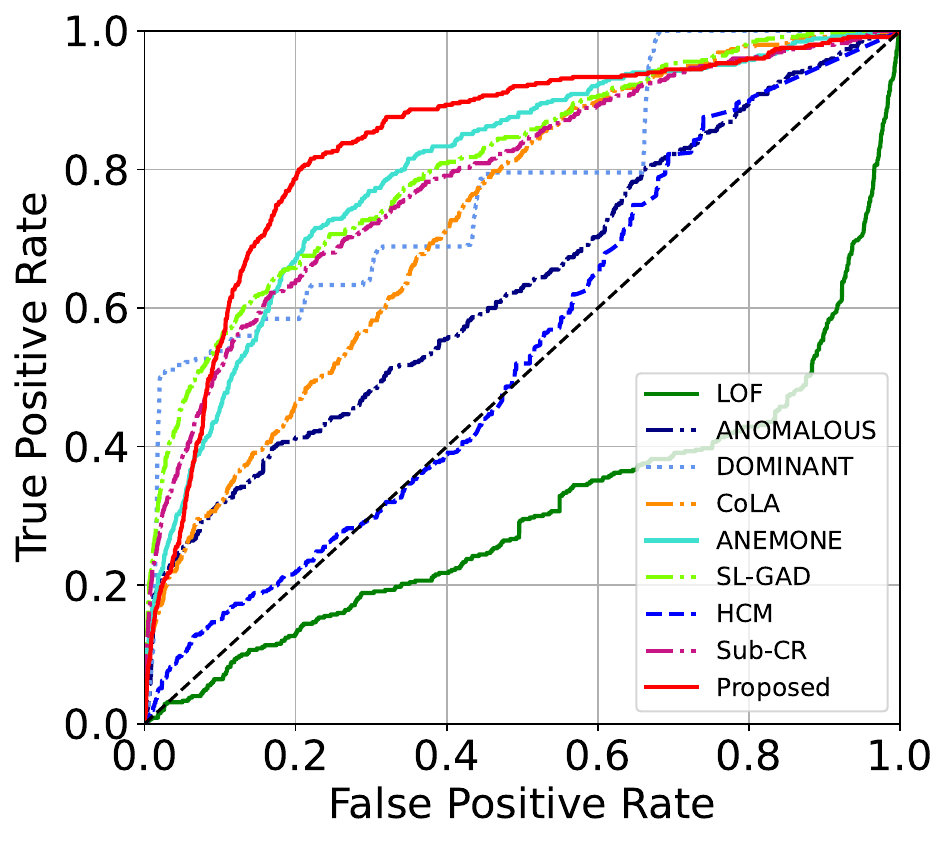}
}
\subfloat[ACM]{
\includegraphics[width=0.31\textwidth]{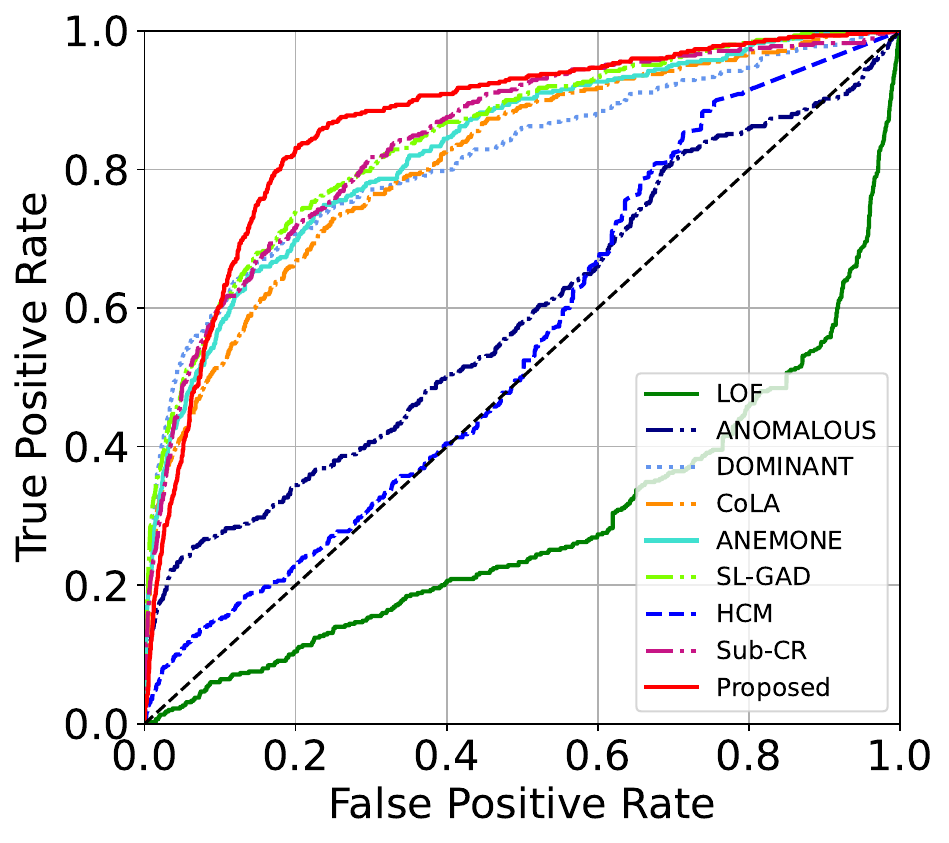}
}
\caption{ROC curves comparison on six benchmark datasets. The area under the curve is larger, the anomaly detection performance is better. The black dotted lines are the ``random line'', indicating the performance under random guessing.}
\label{fig:AUC}
\vspace{-10pt}
\end{figure*}

\subsection{Experimental Results}
Table~\ref{table:AUC} shows the results of performance estimates based on AUC values for nine approaches. Simultaneously, we can observe the ROC curves in Figure~\ref{fig:AUC}, which intuitively demonstrate the model performance by their under-line areas. Through comparison, we can draw the following conclusions: 

\begin{itemize}
\item \textbf{NLGAD outperforms the other models on all datasets.} It achieves significant AUC gains of \textbf{0.94\%}, \textbf{5.89\%}, \textbf{2.09\%}, \textbf{0.63\%}, \textbf{3.40\%}, and \textbf{2.05\%} on Cora, UAI2010, CiteSeer, DBLP, Citation, and ACM, separately. These results verify the effectiveness of NLGAD and its great improvement in detection performance. Especially, NLGAD achieves 1.63\%-5.89\% gains on AUC values against ANEMONE.
\item \textbf{Most deep neural network-based methods work better than shallow methods.} Among them, models adopting the contrastive learning paradigm outperform the others. These indicate that deep methods, especially contrastive learning methods, can process high-dimensional graph datasets better than traditional methods and dig valuable features and structure information from them.
\end{itemize}

\subsection{Ablation Study}
\subsubsection{Hybrid Strategy for Normality Selection.}
To confirm the effectiveness of the proposed hybrid strategy for normality selection, we implement ablation study experiments. Instead of the dynamic strategy, we retain the anomaly degree estimates of all nodes at each step and average them before the percent strategy (termed \textbf{NLGAD-AAS}). Then, we conduct another experiment, only retaining the last step's estimates before the percent strategy (termed \textbf{NLGAD-OLS}). Table~\ref{table:selection} shows that NLGAD works better than NLGAD-AAS and NLGAD-OLS on all datasets, which confirms the effectiveness of the strategy. This further indicates that the detection ability of the initial model is relatively weak, and we should gradually introduce the most reliable estimates to the normality pool. The dynamic strategy meets its requirements for speed control, while the percent strategy further guarantees the quality of pseudo-labels.

\begin{table}[ht]
\centering
\caption{Ablation study of normality selection strategy w.r.t. AUC.}
\resizebox{0.47\textwidth}{!}{
\begin{tabular}{ccccccc}
\toprule
 &$\textbf{Cora}$&$\textbf{UAI2010}$&$\textbf{CiteSeer}$&$\textbf{DBLP}$&$\textbf{Citation}$&$\textbf{ACM}$\\
\midrule
NLGAD-AAS& 0.9272 & 0.9227 & 0.9358  & 0.8397 & 0.8099  & 0.8611  \\
NLGAD-OLS& 0.9031  & 0.8995  & 0.8974  & 0.7793  & 0.7853 & 0.7836 \\
\textbf{NLGAD} & \textbf{0.9286}  & \textbf{0.9320} & \textbf{0.9457}  & \textbf{0.8524}  & \textbf{0.8435}  & \textbf{0.8655}    \\
\bottomrule
\end{tabular}}
\label{table:selection}
\vspace{-15pt}
\end{table}

\subsubsection{Normality Learning.}
Besides, we perform another ablation study experiment to verify the validity of normality learning. We train the backbone networks with the same epochs as the normality selection module but do not perform normality selection and normality learning (termed \textbf{NLGAD-OSP}). In the meantime, we train the model using the same number of total epochs as NLGAD without normality selection and normality learning (termed \textbf{NLGAD-SNP}). The former is used to verify the effectiveness of normality learning. The latter excludes interference from the number of total training epochs. Table~\ref{table:normality} demonstrates that NLGAD outperforms the others on all datasets. The results show that the normality learning-based scheme is effective for GAD.

\begin{table}[ht]
\centering
\caption{Ablation study of normality learning w.r.t. AUC.}
\resizebox{0.47\textwidth}{!}{
\begin{tabular}{ccccccc}
\toprule
 &$\textbf{Cora}$&$\textbf{UAI2010}$&$\textbf{CiteSeer}$&$\textbf{DBLP}$&$\textbf{Citation}$&$\textbf{ACM}$\\
\midrule
NLGAD-OSP& 0.9143 & 0.9143 & 0.9340  & 0.8407 & 0.8233  & 0.8612  \\
NLGAD-SNP& 0.9002  & 0.8980  & 0.8979  & 0.8190  & 0.8176 & 0.8551 \\
\textbf{NLGAD} & \textbf{0.9286}  & \textbf{0.9320} & \textbf{0.9457}  & \textbf{0.8524}  & \textbf{0.8435}  & \textbf{0.8655}    \\
\bottomrule
\end{tabular}}
\label{table:normality}
\vspace{-10pt}
\end{table}

\subsection{Sensibility Analysis}
\subsubsection{Percentage $K$ of Nodes in Normality Selection.}
\label{sensibility_K}
\begin{figure}[ht]
    \centering
    \includegraphics[width = 0.45\textwidth]{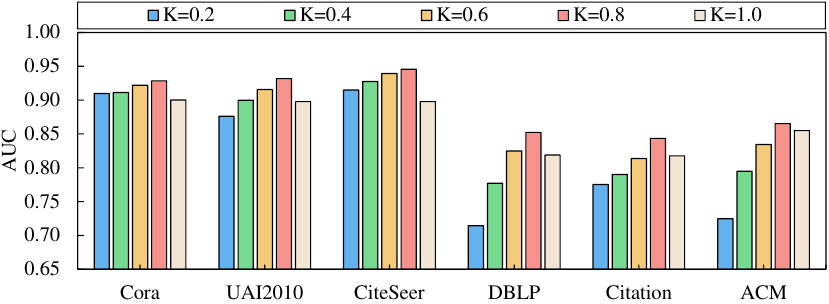}
    \caption{Sensibility analysis of percent strategy ${K}$ w.r.t. AUC.}
    \label{fig:K}
\vspace{-10pt}
\end{figure}

Figure~\ref{fig:K} illustrates the influence of the percentage $K$ of nodes in normality selection on detection performance when $K$ varies from 0.2 to 1.0 with step 0.2. $K=1.0$ means that the model does not perform normality selection. We observe that the performance shows an increasing trend as $K$ increases before $K=1.0$. This shows that normality learning needs a sufficient number of normal nodes as the input. The model cannot learn the pattern of normality well by the deficiency of normal nodes. In the meantime, the normality learning-based scheme can further improve detection performance. We fix $K$ to 0.8 on all datasets.

\begin{figure}[ht]
    \centering
    \includegraphics[width = 0.45\textwidth]{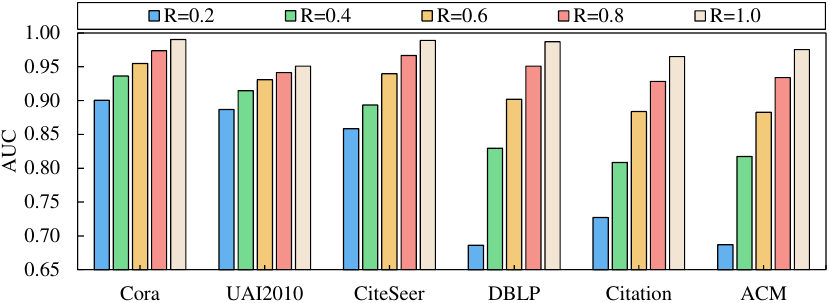}
    \caption{Percentage ${R}$ of normal nodes w.r.t. AUC.}
    \label{fig:R}
\vspace{-10pt}
\end{figure}

And DBLP, Citation, and ACM are more sensitive to $K$ than Cora, UAI2010, and CiteSeer. We conduct additional experiments to further explore the reasons for this phenomenon. We randomly select $R$ percent normal nodes as the training input of the model (without normality selection and learning) when $R$ varies from 0.2 to 1.0. As shown in Figure~\ref{fig:R}, the performances on DBLP, Citation, and ACM are greatly influenced by $R$, which is similar to the impact of $K$. This indicates that the difficulty to learn the pattern of normal nodes is different on different datasets. And such three datasets need more reliable normal nodes than the other datasets in the normality learning phase.

\subsubsection{Number of Steps $T_{s}$ in Normality Selection.}
Then we explore the effect of $T_{s}$ on the performance. Figure~\ref{fig:epoch1} shows $T_{s}$ has different impacts on different datasets. The curve of Citation shows an upward and then downward trend. CiteSeer suffers from little influence of $T_{s}$. The others rise first, then remain essentially fixed. These indicate that too many steps in normality selection can not effectively improve the detection ability but bring useless calculations. To balance performance and complexity, we set $T_{s}=200$ on Cora, UAI2010, and CiteSeer, $T_{s}=300$ on DBLP, Citation, and ACM.

\subsubsection{Number of Epochs $T_{r}$ in Normality Learning.}
\begin{figure}[ht]
\vspace{-10pt}
\centering
\subfloat[Normality Selection Steps $T_{s}$]{
\includegraphics[width=0.25\textwidth]{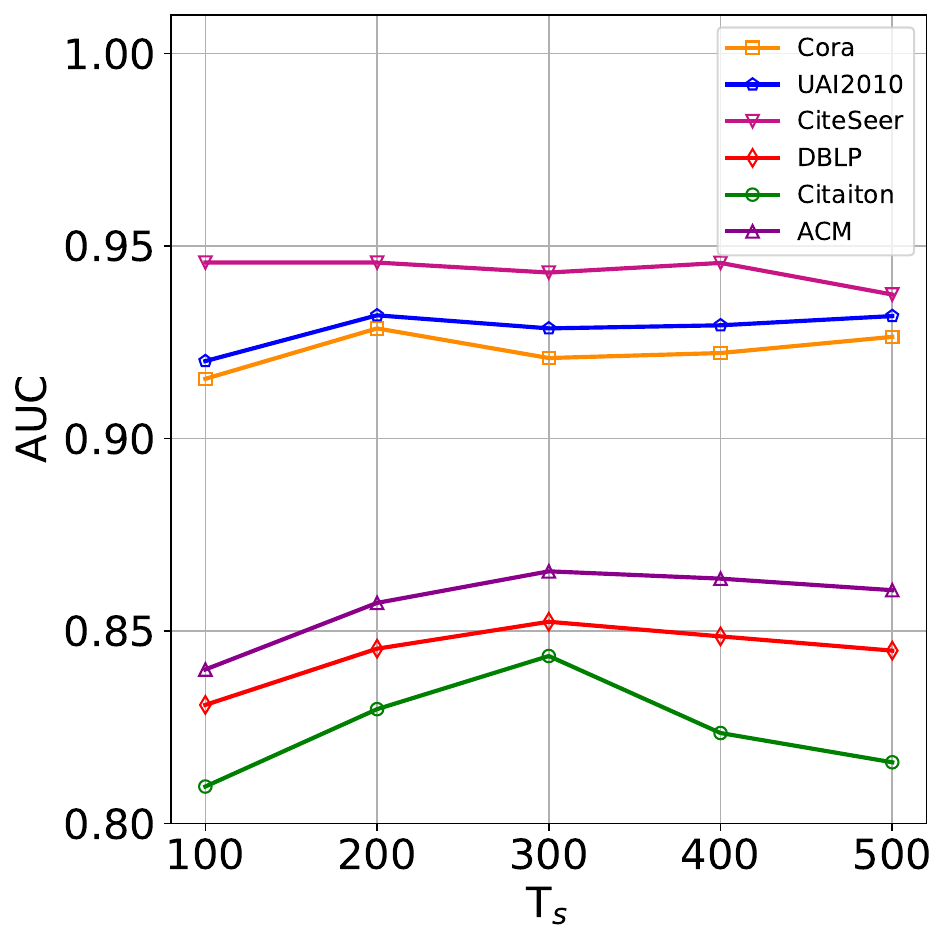}
\label{fig:epoch1}
}
\subfloat[Normality Learning Epochs $T_{r}$]{
\includegraphics[width=0.25\textwidth]{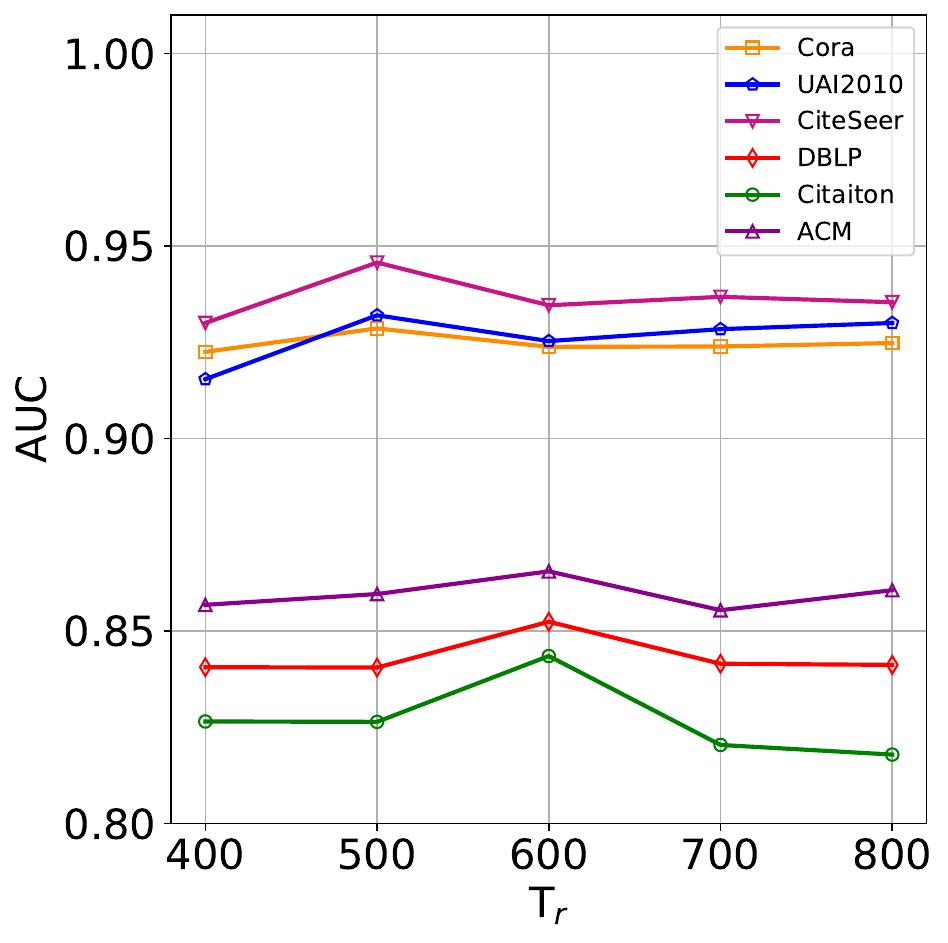}
\label{fig:epoch2}
}
\caption{Sensibility analyses of normality selection steps $T_{s}$ and normality learning epochs $T_{r}$ w.r.t. AUC.}
\vspace{-15pt}
\end{figure}

Figure~\ref{fig:epoch2} shows the impact of $T_{r}$ on the detection performance. The curves on Cora and ACM are less influenced by $T_{r}$. And most curves rise first and then fall. It manifests that the model may suffer from being overfitted on these datasets when $T_{r}$ is set too larger. In practice, we set $T_{r}=500$ on Cora, UAI2010, and CiteSeer, $T_{r}=600$ on DBLP, Citation, and ACM.

\subsubsection{Loss Balance Parameter $\alpha$.}

\begin{figure}[!ht]
    \centering
    \includegraphics[width = 0.35\textwidth]{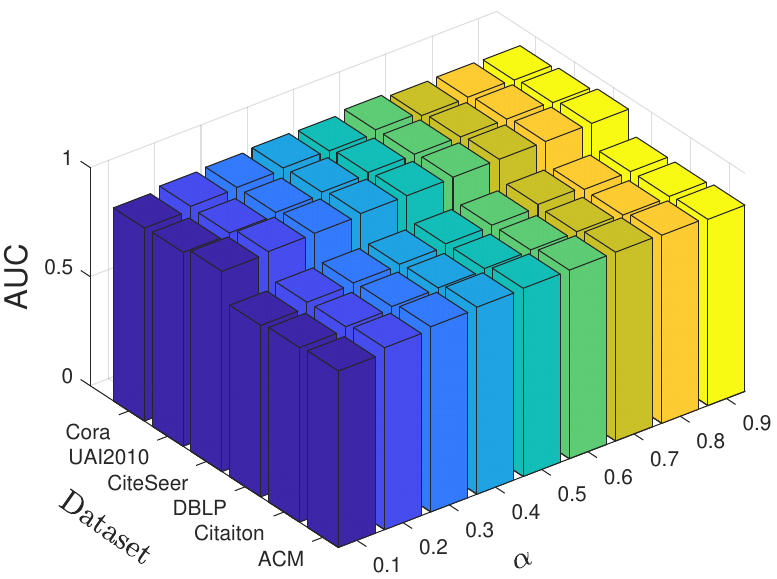}
    \caption{Loss balance parameter ${\alpha}$ w.r.t. AUC values.}
    \label{fig:Alpha}
\vspace{-10pt}
\end{figure}

We discuss the vital balance parameter $\alpha$ in the loss function. As illustrated in Figure~\ref{fig:Alpha}, the influence of $\alpha$ on detection performance shows a first upward and then downward trend on all datasets. The multi-scale contrastive strategy effectively digs different anomalous information, which contributes to the GAD task. In practice, we set $\alpha$ to 0.6, 0.6, 0.9, 0.7, 0.7, and 0.7 on Cora, UAI2010, CiteSeer, DBLP, Citation, and ACM, respectively.

\section{Conclusion}
In this paper, we explain what normality learning is in GAD and how it can help to detect anomalies in graph. Then, we devise a normality learning-based framework, NLGAD. Extensive experiments on six benchmark datasets confirm that NLGAD outperforms the state-of-the-art approaches. In the future, we will further explore the strategy of normality selection that can obtain sufficient and reliable normal nodes.

\begin{acks}
This work was supported by the National Key R\&D Program of China (project no. 2020AAA0107100) and the National Natural Science Foundation of China (project no. 62325604).
\end{acks}

\bibliographystyle{ACM-Reference-Format}
\balance
\bibliography{acmart}









\clearpage
\appendix

\section{Algorithm}
The overall procedures of NLGAD are shown in Algorithm~\ref{ALGORITHM}.

\begin{algorithm}[!h]
\small
\caption{The proposed NLGAD.}
\label{ALGORITHM}
\flushleft{\textbf{Input}: An undirected graph $\mathcal{G} = \left ( \mathcal{V}, \mathcal{E} \right )$; Number of normality selection steps $T_{s}$; Number of refine training epochs $T_{r}$; Batch size $B$.} \\
\flushleft{\textbf{Output}: Anomaly score function $f\left ( \cdot  \right ) $.}
\begin{algorithmic}[1]
\STATE Initialize the model with subgraph-node and node-node contrasts. 
\FOR{$t=1$ to $T_{s}$}
\STATE $\mathcal{V}$ is divided into batches with size $B$ by random.
\FOR{$v_{i} \in B$}
\STATE In each contrast, estimate the similarity of the target node and counterparts in positive and negative pairs.
\STATE Calculate the anomaly degree estimations for each node.
\STATE Select high-confident anomaly degree estimation into the normality pool.
\STATE Calculate the joint loss of subgraph-node and node-node contrasts.
\STATE Back propagation and update trainable parameters.
\ENDFOR
\IF{$t=T_{s}$}
\STATE Average the mean anomaly degree estimations for each node.
Assign $K$ percent nodes with normal pseudo labels.
\ENDIF
\ENDFOR
\FOR{$t=1$ to $T_{r}$}
\STATE Refine the model with the input of selected normal nodes.
\ENDFOR
\STATE By multiple round detections, calculate the final anomaly score for each node.
\end{algorithmic}
\end{algorithm}

\section{Complexity Comparison}
We conduct time complexity analysis of the SOTA GAD models. $T_{s}+T_{r}$ is the total training epochs of NLGAD. And $T$ is the number of epochs for the other models. $r$ is the number of rounds in the inference phase. The other symbol definitions are consistent with Subsection~\ref{complexity}. Table~\ref{table:time} shows that  our method improves performance without significantly increasing complexity.

\begin{table}[!h]
\centering
\caption{Comparison of time complexity for different anomaly detection methods in GAD.}
\resizebox{0.48\textwidth}{!}{
\begin{tabular}{ccc}
\toprule
\textbf{Method}& \textbf{Time Complexity} & \\
\midrule
LOF~\cite{breunig2000lof} &$\mathcal{O}\left (n\log{n} \right )$\\
ANOMALOUS~\cite{peng2018anomalous} &$\mathcal{O}\left (n^{2}d \right )$\\
DOMINANT~\cite{ding2019deep} &$\mathcal{O}\left ( \left (Kmd+Knd^{2} \right ) T \right )$\\
CoLA~\cite{liu2021anomaly} &$\mathcal{O}\left ( n \left (c\delta+Kqd+Kcd^{2}\right ) \left (T+ r \right ) \right )$\\
ANEMONE~\cite{jin2021anemone} &$\mathcal{O}\left ( n \left (c\delta+Kqd+Kcd^{2}\right ) \left (T+ r \right ) \right )$\\
SL-GAD~\cite{zheng2021generative} &$\mathcal{O}\left ( n \left (c\delta+Kqd+Kcd^{2}\right ) \left (T+ r \right ) \right )$\\
HCM~\cite{huang2021hop} &$\mathcal{O}\left ( m + \left ( m+n\right )\log{n} + KmdT \right )$\\
Sub-CR~\cite{zhang2022reconstruction} &$\mathcal{O}\left ( n^{3}+n \left (c\delta+Kqd+Kcd^{2}\right ) \left (T+ r \right ) \right )$\\
\midrule
NLGAD (Proposed) &$\mathcal{O}\left (  n\left (c\delta+Kqd+Kcd^{2}\right )\left (T_{s}+T_{r}+ r\right ) \right )$\\
\bottomrule 
\end{tabular}
}
\label{table:time}
\end{table}

\end{document}